# STAIR Actions: A Video Dataset of Everyday Home Actions

Yuya Yoshikawa, Jiaqing Lin, Akikazu Takeuchi

STAIR Lab, Chiba Institute of Technology

**Abstract.** A new large-scale video dataset for human action recognition, called **STAIR Actions** is introduced. STAIR Actions contains 100 categories of action labels representing fine-grained everyday home actions so that it can be applied to research in various home tasks such as nursing, caring, and security. In STAIR Actions, each video has a single action label. Moreover, for each action category, there are around 1,000 videos that were obtained from YouTube or produced by crowdsource workers. The duration of each video is mostly five to six seconds. The total number of videos is 102,462. We explain how we constructed STAIR Actions and show the characteristics of STAIR Actions compared to existing datasets for human action recognition. Experiments with three major models for action recognition show that STAIR Actions can train large models and achieve good performance. STAIR Actions can be downloaded from `http://actions.stair.center`.

## 1 Introduction

In recent years, human action recognition, the task of classifying what actions people appearing in a given video are performing, has attracted attention as one of the main themes in video analysis [1, 2]. Nowadays, because cameras are installed in many devices such as smartphones, robots, cars, and home appliances, it is expected that the technologies for human action recognition will be used in various situations instead of recognition by human eyes.

In most recent studies on human action recognition, deep neural networks (DNNs) are used. When using DNNs, the following two points are important. The first is the size of the dataset. One reason image recognition has become successful is the massive amount of labeled image data, which is now sufficient for training DNNs, along with the technical revolution of deep learning [3]. Hara et al. suggested this successful scheme could also be valid for human action recognition in videos [4]. The second is the selection of action labels. If one would like to apply the trained model to do some task, action labels should be chosen from that task domain. Although many datasets for human action recognition have been constructed so far, action labels were not designed with target tasks in mind, but instead designed considering the ease of data collection.

In this paper, we introduce a new large-scale video dataset for human action recognition called **STAIR Actions**. STAIR Actions contains 100 categories of action labels representing fine-grained everyday home actions so that it can be used for the recognition of various home tasks such as nursing, caring, and security. For each action category, there are 900 to 1,200 videos obtained from YouTube or produced by crowdsource workers. Moreover, each video in STAIR Actions has a single action label. The duration of each video is mostly 5–6 s. The shortest and longest are 3 and 10 s, respectively. The total number of videos is 102,462. In this paper, we explain how we constructed STAIR Actions, and compare the characteristics of STAIR Actions with existing datasets for human action recognition. Experiments of training three popular architectures show that STAIR Actions can succeed in training large models that achieve competitive performance. STAIR Actions can be downloaded from `http://actions.stair.center`.

Our main contributions can be summarized as follows:

- We developed a large video dataset of everyday home actions consisting of a balanced distribution of videos over 100 action categories.
- Through the analysis of the dataset, we characterized the dataset against existing datasets and identified challenging action categories in this dataset.
- We showed how human action recognition can be achieved with this dataset by experiments using three DNN models for video.

The rest of this paper is organized as follows. In Section 2, we review the existing action recognition datasets widely used in computer vision research. In Section 3, we explain how we constructed STAIR Actions in detail. Next, the unique characteristics of STAIR Actions are shown in Section 4. Benchmark results of training popular architectures with STAIR Actions are described in Section 5. Finally, we conclude this paper in Section 6.

## 2 Related Work

Over the last two decades, several action video datasets have been developed.

In the 2000s, relatively small datasets consisting of tens to hundreds of videos were constructed. For example, the *KTH* dataset consists of 600 monochrome videos with six action categories [5], the *Hollywood* dataset includes 430 short video clips extracted from 32 movies with eight action categories [6], and the *UCF11* dataset contains videos obtained from YouTube with 11 sports-related action categories [7].

Subsequently, datasets have become larger because a large number of videos can easily be gathered from social media such as YouTube. The *HMDB51* dataset includes 6,849 videos and its action labels comprise both indoor and outdoor actions [8]. The *UCF101* dataset is an extension of *UCF50*, which mostly consists of sports actions [9]. UCF101 includes 13,320 videos with 101 action labels, each of which can be classified into one of five large classes: human-object interaction, body motion only, human-human interaction, playing musical instruments, and sports.

Since 2015, several datasets that have a large number of categories and/or videos have been constructed. Version 1.3 of *ActivityNet* contains 23,064 video clips extracted from YouTube videos, each of which is annotated by one of 200 action labels [10]. *Kinetics* contains 400 human action categories with at least 400 videos clips for each action [11]. Each clip lasts around 10 s and is taken from a different YouTube video. The *AVA2.0* dataset provides 80 atomic visual actions densely annotated in 192 15-min video clips, resulting in 740,000 action labels in total [12].

STAIR Actions is a large video dataset for human action recognition that is competitive in size with respect to both videos and action labels with ActivityNet, Kinetics, and AVA. Moreover, the action labels of STAIR Actions are focused and fine-grained; that is, all the videos in STAIR Actions are related only to everyday home actions. In Section 4, we compare the characteristics of STAIR Actions with those of ActivityNet, Kinetics, and AVA, qualitatively and quantitatively.

## 3 Data Collection

This section explains how we construct STAIR Actions dataset in details.

### 3.1 Selection of the 100 Action Labels

This subsection explains how we selected the 100 action labels in STAIR Actions.

First, we obtained the Japanese basic verb list from Wiktionary,[1] which contains 170 basic verbs. From the list, we extracted the verbs associated with actions commonly seen in the home and office. Moreover, we added verbs characteristically associated with special rooms such as the bathroom, kitchen, and living room.

Some verbs change their meanings depending on the object words with which they are used. For example, the verb "open" is related to various actions such as "opening a door," "opening a refrigerator door," and "opening a bottle." Unlike *Moments in Time* dataset, which intentionally adopts verbs as action labels [13], we believe it is important to distinguish such actions, even if they share the same verb. Therefore, we defined action labels using the form "verb + object" for such verbs. The selected 100 action labels are listed in Table 1.

Note that the action labels of Kinetics and ActivityNet seem to be selected from keywords that return a large number of videos on YouTube. Unlike these datasets, those of STAIR Actions are selected in a top-down manner from actions that need to be recognized in the home and office.

[1] `https://en.wiktionary.org/wiki/Appendix:1000_Japanese_basic_words`

**Table 1.** 100 actions of STAIR Actions

| Kitchen related |
|---|
| drinking |
| eating meal |
| eating snack |
| washing dish |
| throwing trash |
| washing hands |
| opening refrig door |
| pouring tea or coffee |
| cutting food |
| cooking |
| **Washroom related** |
| setting hair |
| drying hair with blower |
| making up |
| manicuring |
| gargling |
| brushing teeth |
| washing face |
| shaving |
| **Object manipulation** |
| wearing glass |
| playing with toy |
| playing board game |
| using computer |
| listening to music with headphones |
| playing computer game |
| taking photo |
| using smartphone |
| using tablet |
| operating remote control |
| watching TV |
| telephoning |
| gardening |
| playing guitar |
| playing piano |
| blowing flute |
| standing on chair or table or stepladder |
| throwing |
| opening or closing container |
| smoking |
| ironing |
| knitting or stitching |
| polishing shoe |
| wearing shoes |
| sewing |
| hanging out or capture laundry |
| folding laundry |
| wearing tie |
| putting off cloth |
| putting on cloth |
| housecleaning |
| wiping window |
| drawing picture |
| doing origami |
| reading newspaper |
| studying |
| reading book |
| writing |

| Multiplayer action |
|---|
| changing baby diaper |
| bottle-feeding baby |
| piggybacking someone |
| holding someone |
| feeding baby |
| assisting in getting up |
| assisting in walking |
| teaching |
| nodding |
| shaking head |
| speaking |
| hearing |
| pointing with finger |
| caressing head |
| kissing |
| doing high five |
| hugging |
| stroking animal |
| shaking hands |
| bowing |
| giving massage |
| passing something |
| doing paper-rock-scissors |
| fighting |
| **Solo action** |
| walking with stick |
| walking |
| going up or down stairs |
| jumping on sofa or bed |
| baby crying |
| baby crawling |
| exercising |
| dancing |
| running around |
| clapping hands |
| sitting down |
| standing up |
| sleeping on bed |
| lying on floor |
| leaving room |
| entering room |
| being angry |
| being surprised |
| crying |
| smiling |

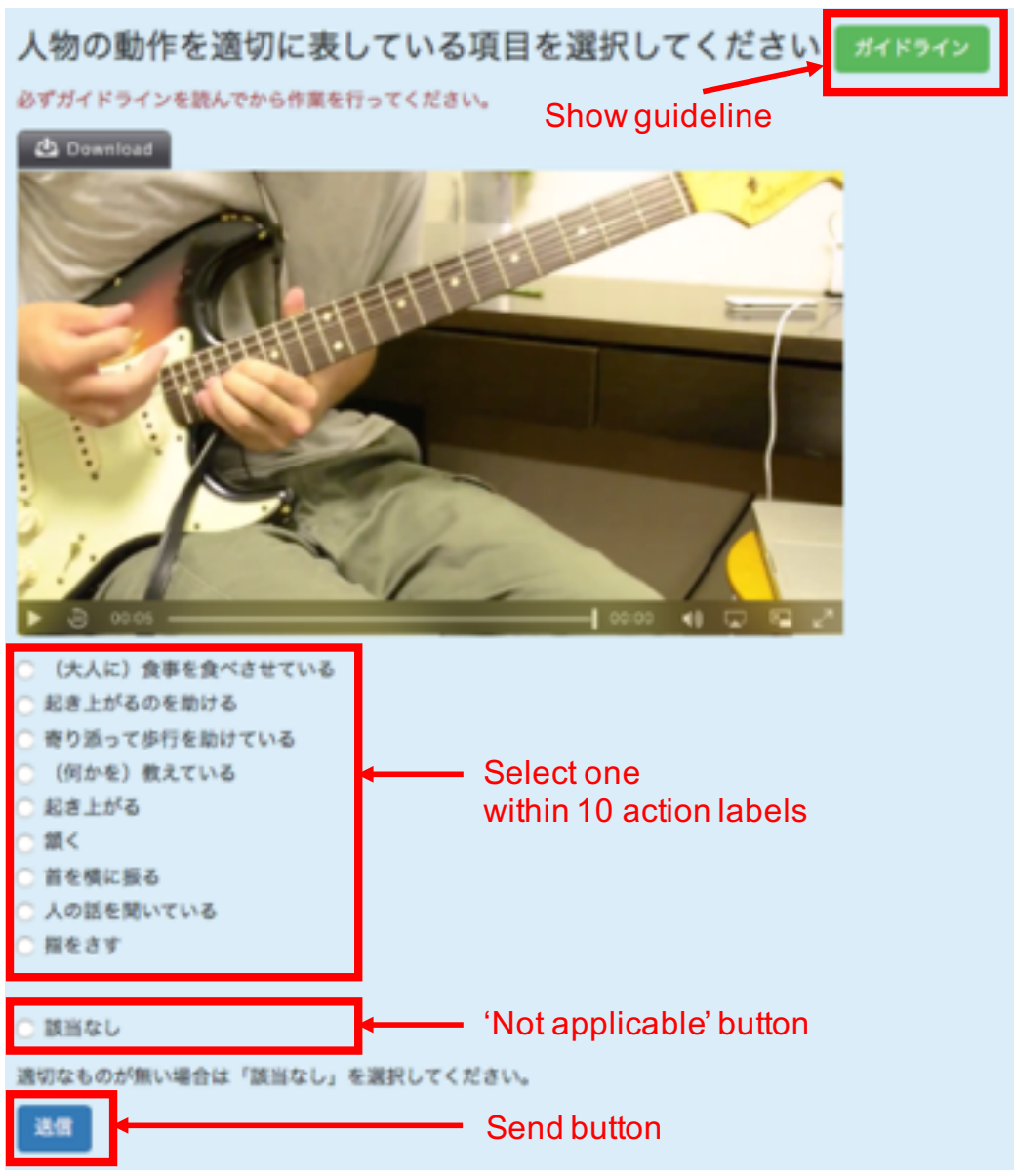

**Fig. 1.** Screenshot of the web annotation system we developed. The system was developed for Japanese people.

### 3.2 Videos from YouTube

About the half of the videos included in STAIR Actions consist of videos from YouTube. We annotated the action labels for the videos using the following four steps:

**Step 1.** Gathering videos from YouTube.
**Step 2.** Extracting 5 s videos from the obtained videos.
**Step 3.** Annotating the 5 s videos with action labels.
**Step 4.** Checking the quality of the annotated labels.

First, we gathered videos from YouTube. When searching for videos, the search keywords were set to words and phrases related to the action labels of STAIR Actions. Additionally, we restricted the YouTube search to videos less than 4 min in duration.

Second, animations and slide-shows were removed from the gathered videos. Scenes with no humans in them were extracted and removed from the videos. The resulting videos were chopped into short video clips 5 s in duration. We believe that 5 s is long enough to recognize an action, while still keeping the input data size manageable.

The third step is to annotate the 5 s video clips with action labels. Annotation was performed by crowdsource workers. The workers were first shown the annotation guidelines and their comprehension of the guidelines was tested. Only the workers who passed the test were asked to annotate videos.

To make the annotation work performed by crowdsource workers efficient, we developed the original web annotation system shown in Figure 1. In the system, the workers are shown a video and asked to select one label from 10 labels plus a "not applicable" label. They were given only 10 labels because workers cannot memorize the guidelines for all 100 actions. A worker is given 10 labels at the beginning of the task and treated as a specialist of those 10 labels.

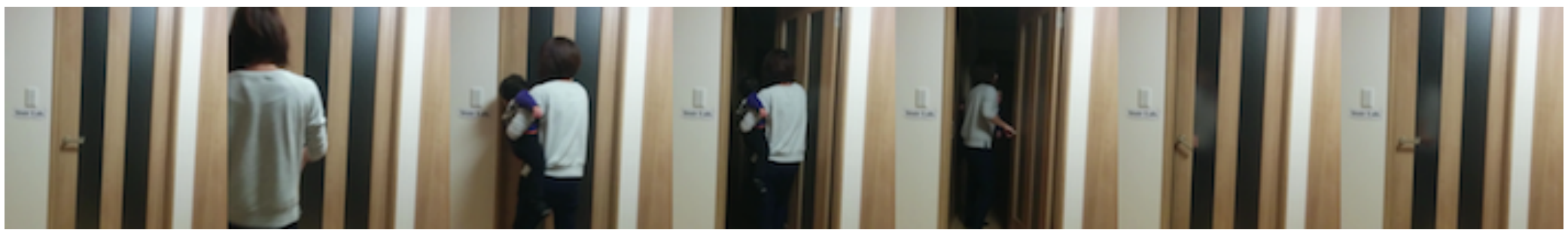

**Fig. 2.** Example of a video for the action "leaving the room."

The final step is to check the quality of the annotated labels by checking whether they are correct. This quality checking was also done by crowdsource workers for all the annotated videos, although these workers were different from those who annotated the videos. In the checking process, each video was checked by three workers, and the final label was added to the dataset by majority vote.

### 3.3 Original Video Collection

For some of the 100 actions in STAIR Actions, it is easy to find corresponding movies on YouTube just by searching using the action labels as keywords. For example, movies of such actions as "cooking," "playing guitar," and "dancing" can be collected rather easily. This is because these action names are popularly used as tags characterizing those movies.

However there are many actions in STAIR Actions that are rarely tagged in YouTube videos. They are ordinary and common actions. Actions such as "entering a room" and "leaving a room" frequently appear in many videos, but nobody tags those scenes using those actions names. Obviously, these actions are not considered worthy of tagging.

To collect videos of those actions, we asked crowdsource workers to take them. Given an action title, a worker was asked to shoot that action at an arbitrary time of day, with any number of actors, and in any place (but preferably in the home). The duration of each video is 5–6 s. There are several risks associated with this method. The biggest one is that someone could send a video extracted from a copyrighted movie. To avoid this, workers are requested to place a paper sign reading "Stair Lab" anywhere in the scene. Figure 2 shows this sign is placed in a scene. Almost half of the videos in STAIR Actions were produced in this way.

Comparing those videos with the YouTube videos, there are several differences.

- There are many videos in which the same person performs the same action. To humans, they look all the same, but pixel-wise, they are different because they were shot under different settings, i.e., at least one of the following conditions differs: camera angle, camera distance, and person's clothing.
- The actions in the videos are more or less staged performances. Actions are always in the center of the scene and the beginning and end of an action are apparent. On the contrary, the actions in YouTube videos are wild. Sometimes it is difficult to see the action.

**Table 2.** STAIR Actions: statistics

| | Training | Validation | Total |
|---|---|---|---|
| # of videos | 92,611 | 9,851 | 102,462 |
| Avg. # of videos per category | 926.11 | 98.51 | – |
| S.D. over categories | 115.02 | 13.23 | – |
| Min–Max | 792–1,108 | 68–128 | – |

**Table 3.** Comparison with existing datasets.

| | # of categories | # of videos | Avg. # of videos per category | # of everyday home actions | Overlap w.r.t. STAIR Actions (SA) |
|---|---|---|---|---|---|
| STAIR Actions | 100 | 102,462 | 1,024 | 100 | – |
| ActivityNet | 203 | 849 h | 137 | 105 | 49 categories match 28 SA categories |
| Kinetics | 400 | 300,000 | 750 | 215 | 122 categories match 54 SA categories |
| AVA | 80 | 57,600 | min=2, max=45,790 | 80 | 40 categories match 43 SA categories |

– The same person performs various actions in same place. Hence, a place is not correlated with the specific action.

## 4 Characteristics of STAIR Actions

In this section, we present the characteristics of STAIR Actions obtained by performing quantitative and qualitative analyses and comparisons with existing datasets.

### 4.1 Statistics

First, we present the basic statistics of STAIR Actions in Table 2. Note that because we split the video clips of STAIR Actions into training and validation sets uniformly at random, the distributions of the number of videos per category for each set are nearly identical.

STAIR Actions focuses on everyday home actions, so all of its action categories are everyday home actions. Existing action datasets have a more diverse interest in various (indoor and outdoor) actions. Table 3 compares four action datasets, STAIR Actions, ActivityNet [10], Kinetics [11], and AVA [12], with respect to the number of everyday home action categories.

We tried to include the Moments in Time database [13] in this table, however, the thinking behind Moments in Time is so different that we could not compare it with the others. There are several radical features in Moments in Time:

– Category = Verb: Each category simply corresponds to a verb. Hence, a verb such as "playing" can contain many action categories such as "playing guitar," "playing a video game," or "playing in the garden," which are regarded as distinct categories in other datasets.

– Non-human actors: Actors in Moments in Time are not limited to humans. The actor could be an animal or a physical object. For example, an actor in the “dropping” category could be a boy, liquid, the jaw of an animated dog, or a boat.

Although Moments in Time seems to be a valuable resource for the exploration of the relationships between language (verb) and vision (motion), we exclude it from this dataset comparison.

As for action vocabulary size, the numbers of categories of the four datasets range between 80 and 400. Because the number of everyday home actions are not given for ActivityNet, Kinetics, and AVA, we counted them ourselves. Based on our analysis, the number of these categories range between 80 to 215.

STAIR Actions, ActivityNet, and AVA have approximately 100 everyday home action categories, respectively, while Kinetics has twice as many such categories. However, there is not a large overlap among them. In the case of STAIR Actions, almost half of its categories are unique and not shared with other datasets. The largest overlap is with Kinetics, which shares 54 categories with STAIR Actions. Indeed, there are many kinds of home actions in our everyday life. Hence, possibility of having the same action category is not very high as long as the size of the video collection is in the several hundreds.

There are many choices that must be made when one defines action categories. For example, one can define “dancing” as one category, but others could decide to define “tango dancing,” “tap dancing,” and “salsa dancing” as separate categories. Hence, the granularity of categories varies from dataset to dataset. Compared with STAIR Actions, ActivityNet and Kinetics have finer-grained categories. Indeed, Kinetics has 18 categories for dancing and ActivityNet has five categories, whilst STAIR Actions has only one category: “dancing.”

As shown in Table 3, 49 categories of ActivityNet match 28 STAIR Actions categories, and 122 categories of Kinetics match 54 STAIR Actions categories. These numbers imply that ActivityNet and Kinetics have categories that are twice as fine as those of STAIR Actions on average.

Fine-grained categories are easy to define because the meaning of the individual category becomes narrower and clearer, hence a classifier may be easier to develop. However, collecting samples of fine-grained categories becomes more difficult because there are more constraints for each sample.

It is now well-known that to obtain better classification accuracy through deep learning, it is critical to have many examples for each category. In the case of ImageNet, that number is 1,000. STAIR Actions and Kinetics are better in this respect. Both provide nearly 1,000 videos for each category. The number of videos per category in AVA varies drastically from category to category. The aim of AVA is on the multiple annotation/labeling of videos, and hence its dataset does not seem to be constructed for classification.

### 4.2 Visually apparent actions

To train a DNN model to recognize a human action, video data should contain at least one human body performing that action. This is practically not so easy from the viewpoint of dataset construction. A typical example is “origami.” Most videos with the “origami” tag on YouTube contain no body and just show the hands. As a compromise, we accepted videos that include only parts of a body.

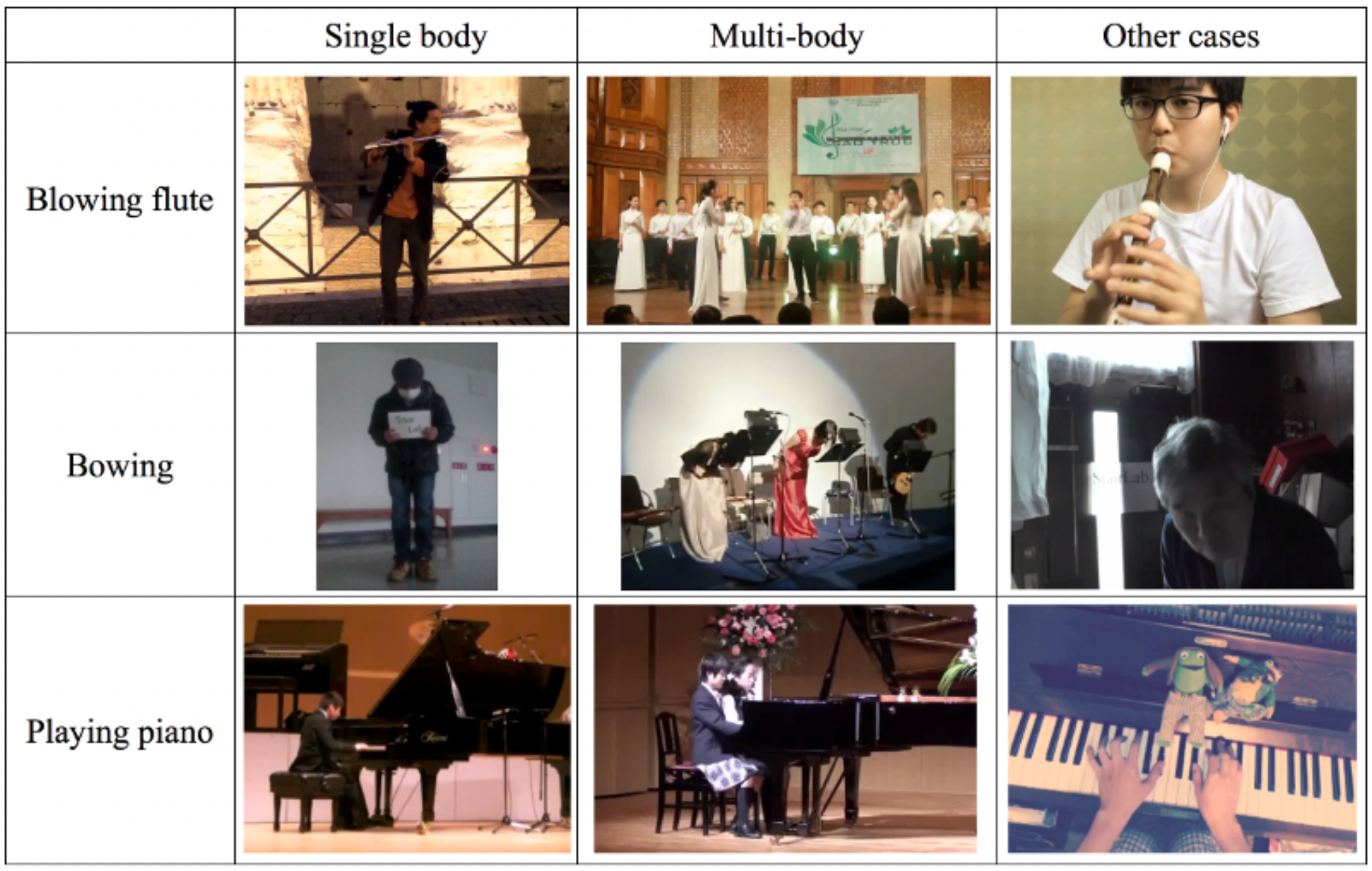

**Fig. 3.** Human parts in different staged scene actions: “Blowing flute,” “Bowing,” and “Playing piano.”

It is interesting to see how many videos in STAIR Actions contain a whole body and how many videos contain just body parts. Hence, we examined each video in the dataset using OpenPose as body/parts detector [14].

The detection procedure was as follows: every tenth frame was extracted from a video, resized to $256 \times 256$, and checked to determine if it includes multiple bodies, a single body, a face, or hands, in this order. If a detection succeeded in the middle of this process, then it became the answer. For example, if a frame contained no body image but a face, then “face” was used as the final answer. After collecting answers from all the sample frames, the result is chosen to be the strongest answer, where the order is defined as follows: multiple bodies $>$ single body $>$ face $>$ hands. Figure 3 shows example frames detected as “single body,” “multi-body,” and other cases (i.e., “face” and “hands”).

Figure 4 shows the result. Although the check is based on tenth-frame sampling, it shows that 39.4% of the videos include at least one frame that includes multiple bodies, and 51.6% of the videos includes a single body, and so on. No body parts are detected only in 4.4% of the videos. Note that more than 91% of the videos in the dataset contain a body image (in the OpenPose sense).

We performed the same body/parts detection check for Kinetics. Figure 5 compares STAIR Actions and Kinetics with respect to body and body parts appearances. Because Kinetics videos are “wild” videos collected from YouTube, OpenPose body/parts detection fails in many cases. This indicates that the STAIR Actions videos contain more human body images than those of Kinetics and may help to train DNNs to recognize actions.

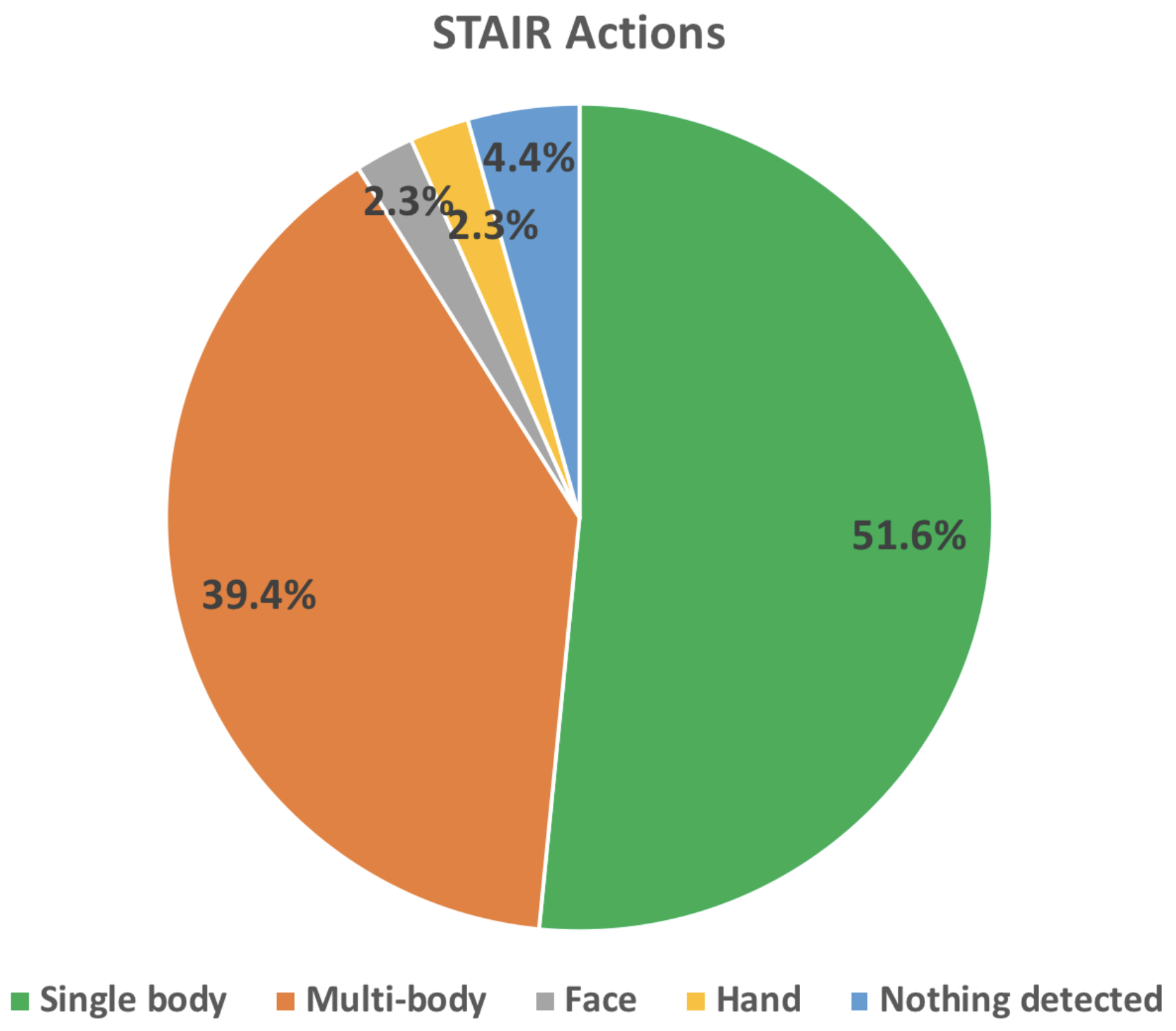

**Fig. 4.** Distribution of human actors/body parts in STAIR Actions videos.

### 4.3 Unique characteristics of STAIR Actions

STAIR Actions is a large video dataset of everyday human actions. Each video included in the dataset is a short video clip 3–10 s in length, mostly 5–6 s in length. Annotation for each video is just one label out of 100 action labels. There is no bounding box, no label taxonomy, and no start and end time.

Moreover, there are the following notable features of STAIR Actions.

**Paired actions where direction matters** STAIR Actions contains paired actions such that the direction of state change matters. Examples are actions such as "sitting down" $\longleftrightarrow$ "standing up," "entering a room" $\longleftrightarrow$ "leaving a room," and "putting on clothing" $\longleftrightarrow$ "taking off clothing". Any static image taken from such paired actions never helps to discriminate which one of the pair is correct. To discriminate actions in the pair, the model needs to recognize the nature of temporal change. This is the core of action recognition.

**Emotional actions** STAIR Actions contains emotional actions such as "being angry," "being surprised," "crying," and "smiling." Those actions are sometimes very subtle and may be difficult to recognize.

**Similar gadgets** To recognize actions related to gadget manipulation, it is important to recognize a gadget in the scene. In STAIR Actions, there are several gadget manipulation categories such as "using smartphone," "taking photo,"

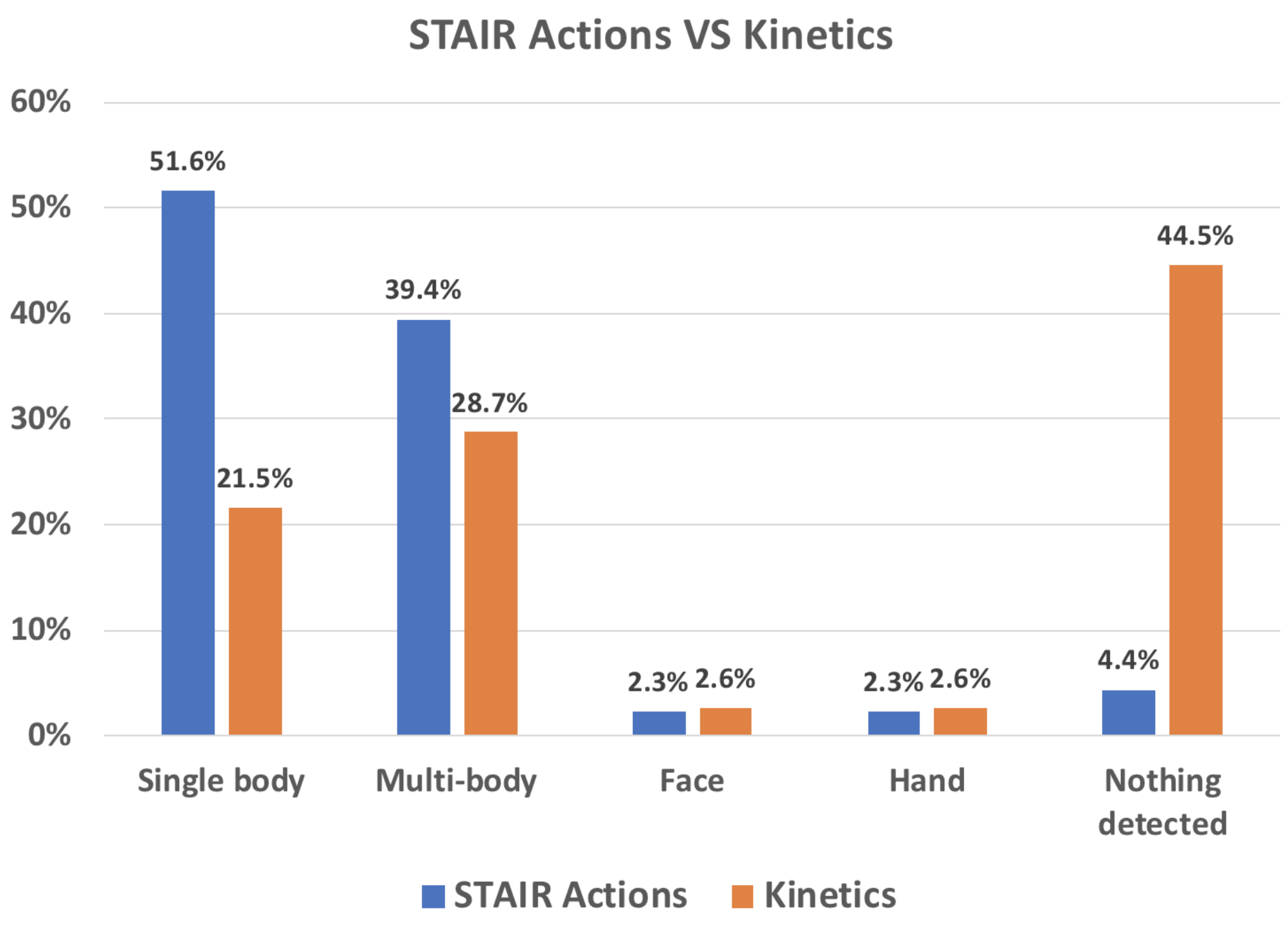

**Fig. 5.** Comparison of the distribution of human actors/body parts in STAIR Actions and Kinetics videos. OpenPose was used as human body/parts detector.

"using tablet," and "operating remote control." Because these devices look similar in many cases, distinguishing such actions may be difficult.

## 5 Action Recognition Benchmarks

In most recent studies on human action recognition, DNNs are used. Such DNN-based models are best trained using a massive number of videos with the correct action labels. DNN-based models can be roughly classified into three types of architectures. The first one is a 2DCNN+LSTN, which first extracts features from each frame in a video using a two-dimensional (2D) convolutional neural network (CNN), feeds the sequence of the features into an LSTM (recurrent neural network) [15], and finally recognizes an action. The second one is a 3DCNN, which extends the 2DCNNs along the space and time axes [16]. The third one is called a two-stream CNN, which combines different 2DCNNs that capture spatial and temporal structures in videos [4, 17].

### 5.1 2DCNN + LSTM

A video can be seen as a sequence of images. Donahue et al. proposed a model called the *LRCN* for action recognition, which (1) extracts image features of each frame from an input video using a 2DCNN, and then (2) extracts temporal features from the sequence of the image features using an LSTM [15].

We implemented LRCN from scratch, referring to their paper. For the 2DCNN in our implementation, we used AlexNet [18] pretrained on the ImageNet dataset (ILSVRC2012). As preprocessing, each frame of an input video was resized to 256 × 256 and then randomly cropped to a 224 × 224 frame. Image features were extracted from each frame by AlexNet from the randomly selected consecutive 30 frames. Here, the image features are the output of the sixth fully-connected layer (i.e., fc6) in AlexNet. Then, the 30 image features were fed into a single-layer LSTM with 256 hidden units. After all the image features were fed into the LSTM, the output of the LSTM was transformed into the probabilities of the action labels by applying a fully-connected layer.

### 5.2 Two-stream CNN

Simonyan et al. proposed the two-stream convolutional neural networks (CNN) model for action recognition [16]. The model consists of two ConvNet architectures, one for spatial information and the other for temporal information. The softmax scores of the two streams are averaged and used for training a multi-class linear SVM [19].

We implemented the two-stream model from scratch in reference to the paper [16]. As for the spatial model, the layer configuration is same as that of [16]. We pretrained the spatial model on ILSVRC2012. In every iteration, we randomly selected a frame from an input video and resized it to 256 × 256, randomly cropping it to 224 × 224, and then feeding it to the spatial model.

As for the temporal model, we computed optical flow using [20]. We selected every two adjacent frames to generate optical flow images with horizontal and vertical components, respectively. The input to the temporal model consisted of two components of 10 stacked optical flow images: 224 × 224 × 2 × 10. More details are described in [16].

To train the two-stream model, we set the mini-batch size to 32 and set the initial learning rate to 0.001 in both streams. Because the spatial model was pretrained, only the last layer was trained. The training result of the spatial model is shown in Figure 6(a). The learning rate was multiplied by 0.1 at epochs 140 and 190. We trained the temporal model from scratch. The learning rate was multiplied by 0.1 at epoch 330. The training result of the temporal model is shown in Figure 6(b). We used Chainer V3 [21] on a server with eight GeForce GTX 1080Ti GPUs.

Table 4 shows the results of four architectures, the spatial model, temporal model, two-stream model that averages the scores of two streams, and two-stream model that trains an SVM with two stream scores as input, on three datasets: HMDB51, UCF101, and STAIR Actions.

### 5.3 3DCNN

3D CNNs with spatio-temporal 3D kernels (3DCNNs) have been expected to be suitable for action recognition from video. However, because of their large number of parameters, it has been difficult to train them without overfitting.

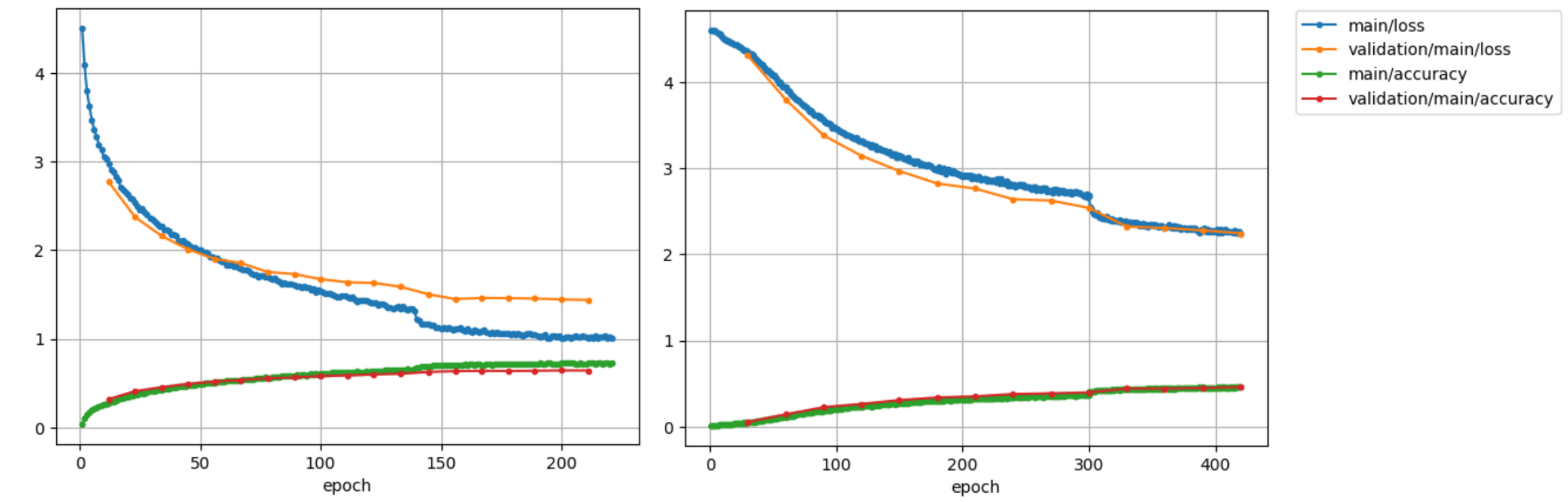

(a) (b)

**Fig. 6.** Performance the two-stream CNN on STAIR Actions on. (a) Result trained on the spatial CNN. (b) Result trained on the temporal CNN.

**Table 4.** Two-stream model accuracies on HMDB51, UCF101, and STAIR Actions.

| | Spatial | Temporal | Avg. fusion | SVM fusion |
|---|---|---|---|---|
| HMDB51 | 40.5% | 54.6% | 58.0% | 59.4% |
| UCF101 | 73.0% | 83.7% | 86.9% | 88.0% |
| STAIR Actions | 70.4% | 54.1% | 73.1% | 73.7% |

Kay et al. [11] and Hara et al. [4, 17] demonstrated that using a large video dataset such as Kinetics, 3DCNN can be trained without falling into overfitting and achieve competitive performance with other modern models such as the two-stream models.

We experimented with a 3DCNN model called ResNet-34, which developed by Hara et al. [2] Although ResNet-34 is neither the deepest nor the best performing model of their 3DCNN models, its performance is promising. The result is shown in Table 6. Most of the model parameters are the same as the one described in [4, 17] except for the number of categories and sample duration, which is the number of consecutive frames to be processed. We experimented using 16, 30, and 60 frames for sample duration. The result is shown in Table 5. There are advantages and disadvantages with respect to the length of sample duration; shorter durations risk missing critical moments of action while longer durations may include irrelevant actions. In addition, longer durations require more parameters to be trained. From the results in Table 5, a duration 30 frames seems good compromise. Figure 7 shows validation loss and accuracy over 200 epochs.

Table 6 summarizes the accuracies of the three models on the three datasets.[3] It is interesting to compare the performances of the same 3DCNN (i.e., ResNet-34) on STAIR Actions and Kinetics. [4] reports that ResNet-34 on Kinetics achieved 60.1% top-1 accuracy. The better performance (76.5%) of the same model on STAIR Actions may be because STAIR Actions has smaller number of categories than Kinetics and its videos contain more body/parts images. Moreover, using deeper models such as ResNeXt-101 or Wide ResNet-50 may make it possible to achieve better accuracy on STAIR Actions.

[2] `https://github.com/kenshohara/3D-ResNets-PyTorch`

[3] The accuracy of LRCN on STAIR Actions will be provided in a later version of the paper.

**Table 5.** 3DResNet-34: Comparison of sample duration at 200 epochs.

| Sample duration (frames) | Loss | Top-1 Accuracy |
|---|---|---|
| 16 | 1.0959 | 0.7387 |
| 30 | 0.9816 | 0.7646 |
| 60 | 1.1717 | 0.7306 |

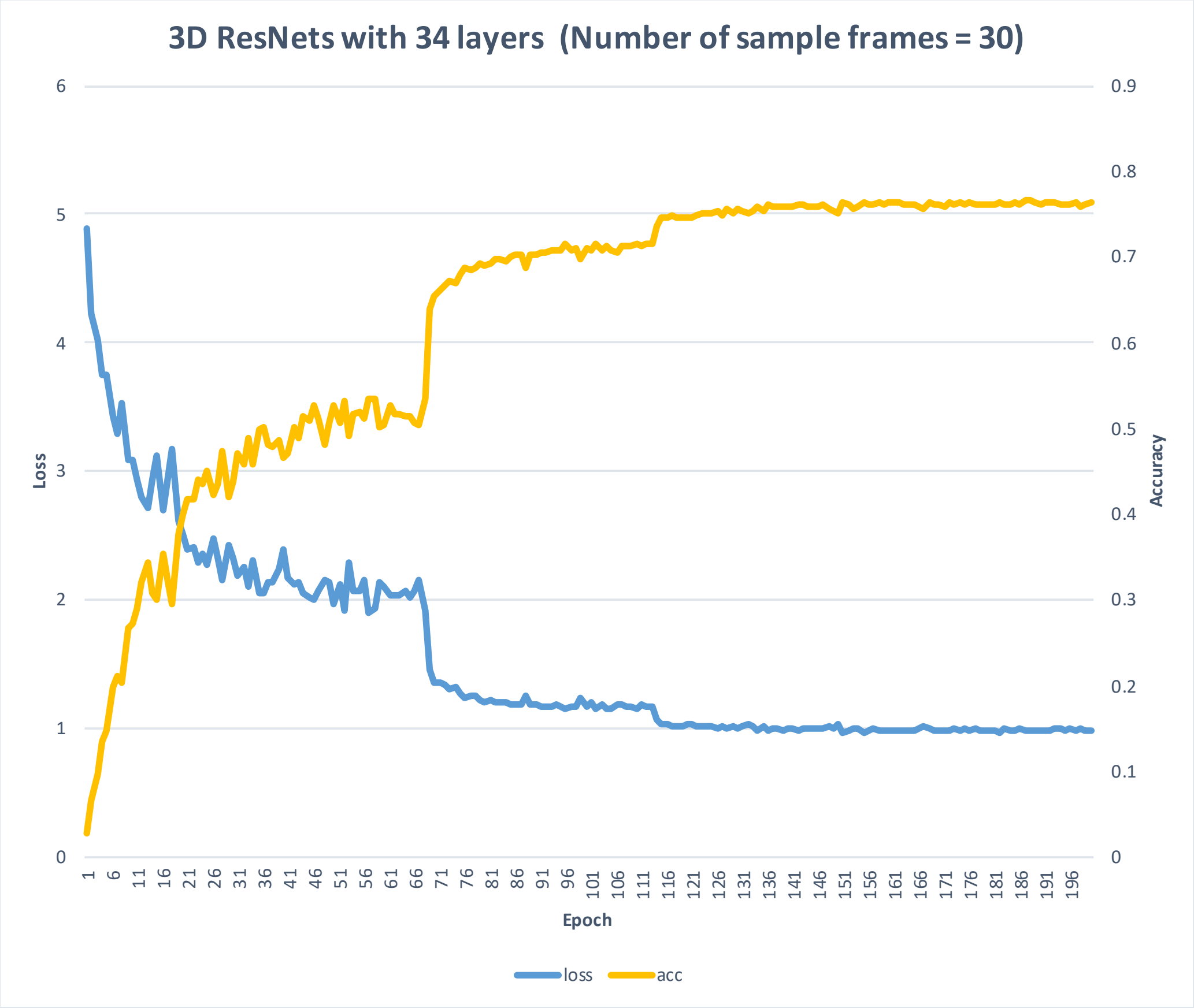

**Fig. 7.** 3D ResNet Performance

## 6 Conclusion

A new video dataset of everyday human actions, STAIR Actions, was introduced. It contains 100 everyday human action categories with an average of 1,000 trimmed video clips for each category. Clips were taken from YouTube or created by crowdsource workers.

STAIR Actions is the first large video dataset of everyday human actions with a balanced distribution of videos over categories. Through experiments with well-known actions recognition models, STAIR Actions is shown to be able to train large models such as 3DCNNs.

**Table 6.** Accuracy of various models on HMDB51, UCF101, and STAIR Actions. The 3DCNN scores for HMDB51 and UCF101 are taken from Hara et al. [4], which used a Kinetics-pretrained 3D ResNet-34. The 3DCNN score for STAIR Actions is our result for 30 sample frames.

| | LRCN | Two-stream CNN | 3DCNN |
|---|---|---|---|
| HMDB51 | 53.3% | 59.4% | 59.1% |
| UCF101 | 87.2% | 88.0% | 87.7% |
| STAIR Actions | N/A | 73.7% | 76.5% |

As for future work, we plan to continuously publish subsequent versions of STAIR Actions with some cleaning and fine tuning. In addition, we are developing a Japanese caption dataset for STAIR Actions that will be published in the future. Further experiments with sophisticated models should be performed. These models include I3D [22] and 3D ResNeXt-101 [4]. It would also be interesting to study how the models pretrained on STAIR Actions perform on UCF-101 and HMDB-51.

## Acknowledgement

This project is supported by NEDO (New Energy and Industrial Technology Development Organization), Japan. We would like to thank Yutaro Shigeto for his helpful suggestions and comments.

## References

1. Cheng, G., Wan, Y., Saudagar, A.N., Namuduri, K., Buckles, B.P.: Advances in Human Action Recognition: A Survey. (2015) 1–30
2. Wu, D., Sharma, N., Blumenstein, M.: Recent Advances in Video-Based Human Action Recognition Using Deep Learning: A Review. In: 2017 International Joint Conference on Neural Networks (IJCNN), IEEE (may 2017) 2865–2872
3. Lecun, Y., Bengio, Y., Hinton, G.: Deep learning. Nature **521**(7553) (2015) 436–444
4. Hara, K., Kataoka, H., Satoh, Y.: Can spatiotemporal 3D CNNs retrace the history of 2D CNNs and ImageNet? arXiv preprint **arXiv:1711.09577** (2017)
5. Schuldt, C., Laptev, I., Caputo, B.: Recognizing Human Actions: A Local SVM Approach. In: Proceedings of the Pattern Recognition, 17th International Conference on (ICPR'04) Volume 3 - Volume 03, IEEE Computer Society (2004) 32–36
6. Laptev, I., Marszałek, M., Schmid, C., Rozenfeld, B.: Learning Realistic Human Actions from Movies. 26th IEEE Conference on Computer Vision and Pattern Recognition (2008)
7. Jingen, L., Jiebo, L., Mubarak, S.: Recognizing Realistic Actions from Videos "in the Wild". In: IEEE International Conference on Computer Vision and Pattern Recognition. (2009)
8. Kuehne, H., Jhuang, H., Garrote, E., Poggio, T., Serre, T.: HMDB: A Large Video Database for Human Motion Recognition. In: International Conference on Computer Vision, IEEE (nov 2011) 2556–2563
9. Soomro, K., Zamir, A.R., Shah, M.: UCF101: A Dataset of 101 Human Actions Classes from Videos in the Wild. Technical Report November (2012)
10. Fabian Caba Heilbron, Victor Escorcia, B.G., Niebles, J.C.: Activitynet: A large-scale video benchmark for human activity understanding. In: Proceedings of the IEEE Conference on Computer Vision and Pattern Recognition. (2015) 961–970
11. Kay, W., Carreira, J., Simonyan, K., Zhang, B., Hillier, C., Vijayanarasimhan, S., Viola, F., Green, T., Back, T., Natsev, P., Suleyman, M., Zisserman, A.: The kinetics human action video dataset. CoRR **abs/1705.06950** (2017)
12. Gu, C., Sun, C., Vijayanarasimhan, S., Pantofaru, C., Ross, D.A., Toderici, G., Li, Y., Ricco, S., Sukthankar, R., Schmid, C., Malik, J.: AVA: A video dataset of spatio-temporally localized atomic visual actions. CoRR **abs/1705.08421** (2017)
13. Monfort, M., Zhou, B., Bargal, S.A., Yan, T., Andonian, A., Ramakrishnan, K., Brown, L., Fan, Q., Gutfruend, D., Vondrick, C., et al.: Moments in time dataset: one million videos for event understanding
14. Cao, Z., Simon, T., Wei, S.E., Sheikh, Y.: Realtime multi-person 2d pose estimation using part affinity fields. In: CVPR. (2017)
15. Donahue, J., Hendricks, L.A., Guadarrama, S., Rohrbach, M., Venugopalan, S., Saenko, K., Darrell, T., Austin, U.T., Lowell, U., Berkeley, U.C.: Long-term Recurrent Convolutional Networks for Visual Recognition and Description. In: IEEE Conference on Computer Vision and Pattern Recognition. (2015)
16. Simonyan, K., Zisserman, A.: Two-stream convolutional networks for action recognition in videos. CoRR **abs/1406.2199** (2014)
17. Hara, K., Kataoka, H., Satoh, Y.: Learning spatio-temporal features with 3D residual networks for action recognition. Proceedings of the ICCV Workshop on Action, Gensture, and Emotion Recognition (2017)
18. Krizhevsky, A., Sutskever, I., Hinton, G.E.: ImageNet Classification with Deep Convolutional Neural Networks. Advances In Neural Information Processing Systems (2012) 1–9

19. Crammer, K., Singer, Y.: On the algorithmic implementation of multiclass kernel-based vector machines. Journal of machine learning research **2**(Dec) (2001) 265–292
20. Brox, T., Bruhn, A., Papenberg, N., Weickert, J.: High accuracy optical flow estimation based on a theory for warping. In: European conference on computer vision, Springer (2004) 25–36
21. Tokui, S., Oono, K., Hido, S., Clayton, J.: Chainer: a next-generation open source framework for deep learning. In: Proceedings of Workshop on Machine Learning Systems (LearningSys) in The Twenty-ninth Annual Conference on Neural Information Processing Systems (NIPS). (2015)
22. Carreira, J., Zisserman, A.: Quo vadis, action recognition? A new model and the kinetics dataset. CoRR **abs/1705.07750** (2017)